\documentclass[runningheads]{llncs}
\usepackage{graphicx}
\usepackage{amsmath}
\usepackage{siunitx}
\usepackage{wrapfig}
\usepackage{xspace}
\usepackage[final]{microtype}
\usepackage{booktabs}
\usepackage{appendix}
\usepackage{tabularx}
\usepackage{csquotes}
\usepackage[T1]{fontenc}
\newcolumntype{Y}{>{\centering\arraybackslash}X}
\usepackage{hyperref}
\usepackage[capitalize]{cleveref}
\usepackage[disable]{todonotes}

\newcommand{\Chen}[1]{\todo[linecolor=green,backgroundcolor=green!25,bordercolor=green,inline]{Chen: #1}}

\usepackage[frozencache=true,cachedir=_minted-output]{minted} % Code highlighting
\newcommand{\ie}{\textit{i.e.}\xspace}
\newcommand{\eg}{\textit{e.g.}\xspace}
\newrobustcmd{\B}{\bfseries}
\DeclareMathOperator*{\argmax}{arg\,max}
\DeclareMathOperator*{\argmin}{arg\,min}

\begin{document}
\title{Web Image Context Extraction with
Graph Neural Networks and Sentence Embeddings\\
on the DOM tree}
\titlerunning{WICE with GNNs and embeddings on the DOM tree}
\author{Chen Dang\inst{1}%\orcidID{0000-1111-2222-3333}
\and
Hicham Randrianarivo\inst{1}%\orcidID{1111-2222-3333-4444}
\and\\
Raphaël Fournier-S'niehotta\inst{2}%\orcidID{2222--3333-4444-5555}
\and
Nicolas Audebert\inst{2}%\orcidID{3333--4444-5555-6666}
}%
\authorrunning{C. Dang et al.}
\institute{Qwant SAS, Paris, France\\
\email{chdangg@gmail.com, h.randrianarivo@qwant.com}\and
CEDRIC (EA4629), CNAM Paris, HESAM Université, France\\
\email{fournier@cnam.fr, nicolas.audebert@cnam.fr}\\
}
\maketitle              % typeset the header of the contribution
\begin{abstract}
Web Image Context Extraction (WICE) consists in obtaining the textual information describing an image using the content of the surrounding webpage. A common preprocessing step before performing WICE is to render the content of the webpage.
When done at a large scale (\textit{e.g.}, for search engine indexation), it may become very computationally costly (up to several seconds per page).
To avoid this cost, we introduce a novel WICE approach that combines Graph Neural Networks (GNNs) and Natural Language Processing models.
Our method relies on a graph model containing both node types and text as features. The model is fed through several blocks of GNNs to extract the textual context.
Since no labeled WICE dataset with ground truth exists, %
we train and evaluate the GNNs on a proxy task that consists in finding the semantically closest text to the image caption.
We then interpret importance weights to find the most relevant text nodes and define them as the image context.
Thanks to GNNs, our model is able to encode both structural and semantic information from the webpage.
We show that our approach gives promising results to help address the large-scale WICE problem using only HTML data.

\keywords{Web Image Context Extraction \and Information Retrieval \and Graph Neural Network \and Natural Language Processing}
\end{abstract}
\section{Introduction}

\begin{wrapfigure}{L}{0.5\textwidth}
\vspace{-1.5\intextsep} % Hacky and ugly: TODO: find out where this whitespace comes from
\begin{center}
    \includegraphics[width=0.5\textwidth]{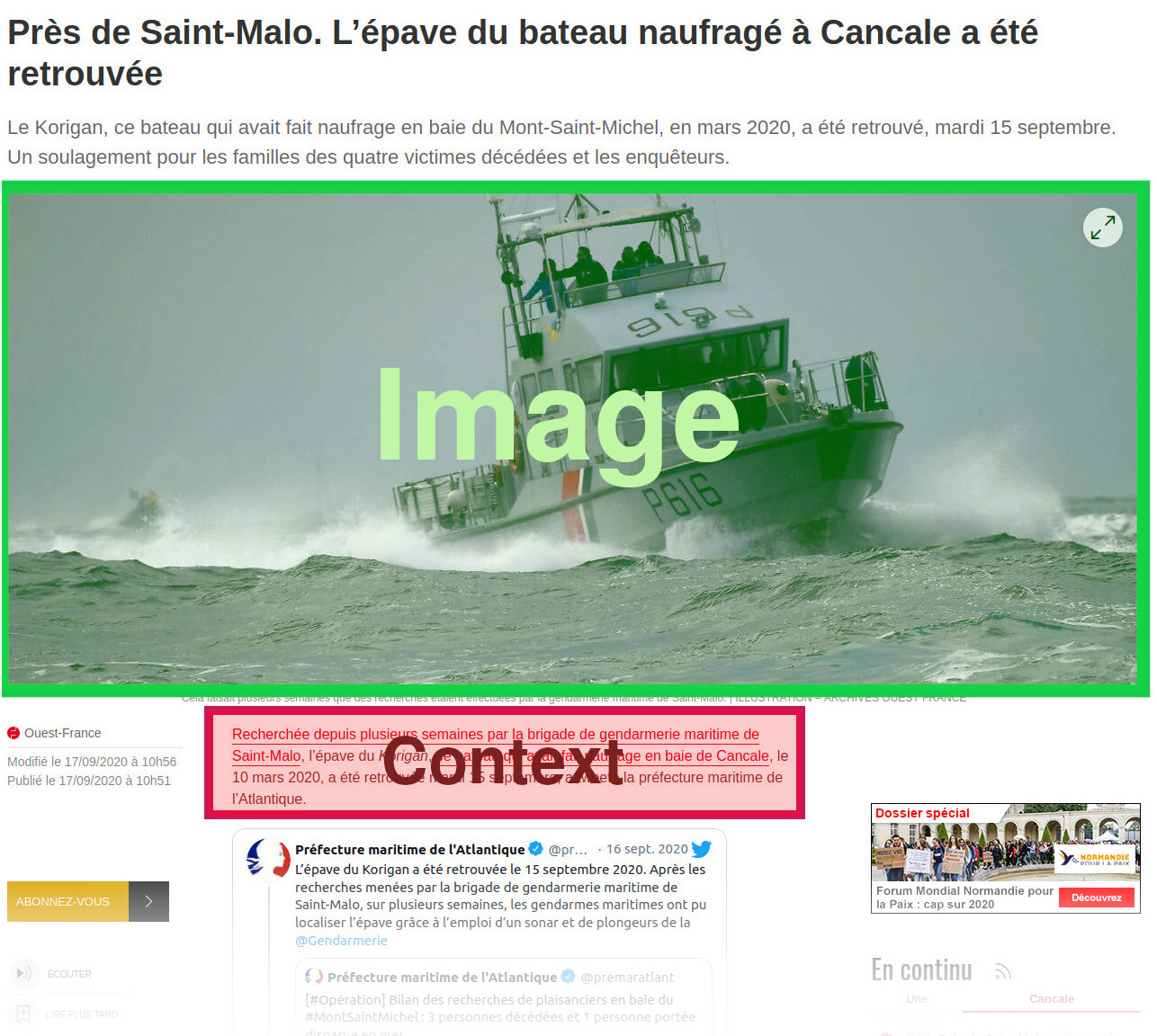}
\end{center}
\vspace{-\intextsep}
\caption{An example of the WICE setting: an image, and its textual description ({\em context}) in the webpage.}
\label{wrap-fig:WICE_illustration}
\vspace{-\intextsep}
\end{wrapfigure}

Searching for images on the Web is essential for Internet users.
Then, there is a need for efficient indexing methods to process large quantities of images.
A useful step for the indexation in an image retrieval system consists in identifying the part of a webpage's text that best describes the image.
The problem of extracting this {\itshape context} from the webpage is called Web Image Context Extraction (WICE) cf. \cref{wrap-fig:WICE_illustration}.
Visually rendering the webpage facilitates the extraction of an image's context, by loading and placing all structural elements of the page at the cost of evaluating several scripts.
On a large scale, visual rendering and content extraction from a webpage is not tractable.
We investigate how the HTML data structure may help in extracting images' contexts.

Many approaches to WICE have been proposed.
\cite{Heng2000} use metadata related to the image as the textual context.
Defining the context as the text in a \textquote{window surrounding the image} in the HTML is common, and some works try to find an optimal number of words to extract around the image \cite{SOUZA2004,FENG2004}.
\cite{GONG2006} consider multiple sources of text, \eg, title and meta information, as the context.
These text-based methods often result in incomplete sentences, and do not provide accurate context when the context and image are not close in the HTML file.

Structure-based approaches focus more on the structure of the HTML document.
An HTML document can be describe as a tree structure where each tag or text an object and the nested objects are \textquote{children} of the enclosing one wich is called the Document Object Model (DOM) \cite{wood1998document}.
Relying on the DOM tree,
some works~\cite{Hattori2007,JOSHI2009} measure similarities between the alternative text of the image and other texts, or develop precise webpage segmentation rules.
\cite{Jia2006} propose a {\em broadcast model} which combines the text blocks around images and information from other webpages linked to the webpage.
\cite{Fauzi2009} classify webpage's structures into three categories and handcraft rules to extract context.
With a high focus on the page structure, DOM-tree based approaches often ignore or fail to fully use textual content, and recent evolutions of webpage programming have rendered many methods based on hard-coded rules no longer applicable.

Finally, some approaches focus on the webpage's visual layout.
\cite{cai2003vips} use visual information to perform the segmentation using a set of predefined rules.
\cite{Tsapatsoulis2016} propose to extract all text which includes a caption and an alternative text at the same level as the image in the DOM tree. They also keep the texts around the image within a radius of \num{0.3} of the webpage rendering height.  \cite{2012_tryfou_extraction} compare visual and semantic clustering and find that visual based clustering performs much better in extracting information of web images

Besides, many graph-based information retrieval methods have been recently proposed.
\cite{qian2019graphie} use a graph-based framework to capture non-local and non-sequential context in sets of sentences.
\cite{de-cao-etal-2019-question} introduce a Graph Neural Network model for multi-step reasoning.
\cite{lockard-etal-2019-openceres} study relation extraction for semi-structured websites.
\cite{2012_tryfou_extraction} compares visual and semantic clustering and find that visual based clustering performs much better in extracting information of web images.

However, browser-based rendering is costly due to CSS and Javascript processing. It is nonetheless required by the state-of-the-art visual WICE approaches that rely on the full styled layout.
We model webpages as graphs whose features are node types and text content.
More specifically, our contributions to the problem of large-scale WICE are:
\begin{itemize}
    \item Inject semantics into the DOM tree using state-of-the-art language models to generate sentence embeddings for each text node,
    \item Model webpages as graphs and use sentence embeddings as node features to train a graph neural network combining structural and semantic information,
    \item Use graphical models for large-scale processing of highly diverse news websites.
\end{itemize}

\begin{figure}[t]
    \centering
    \includegraphics[width=0.9\textwidth]{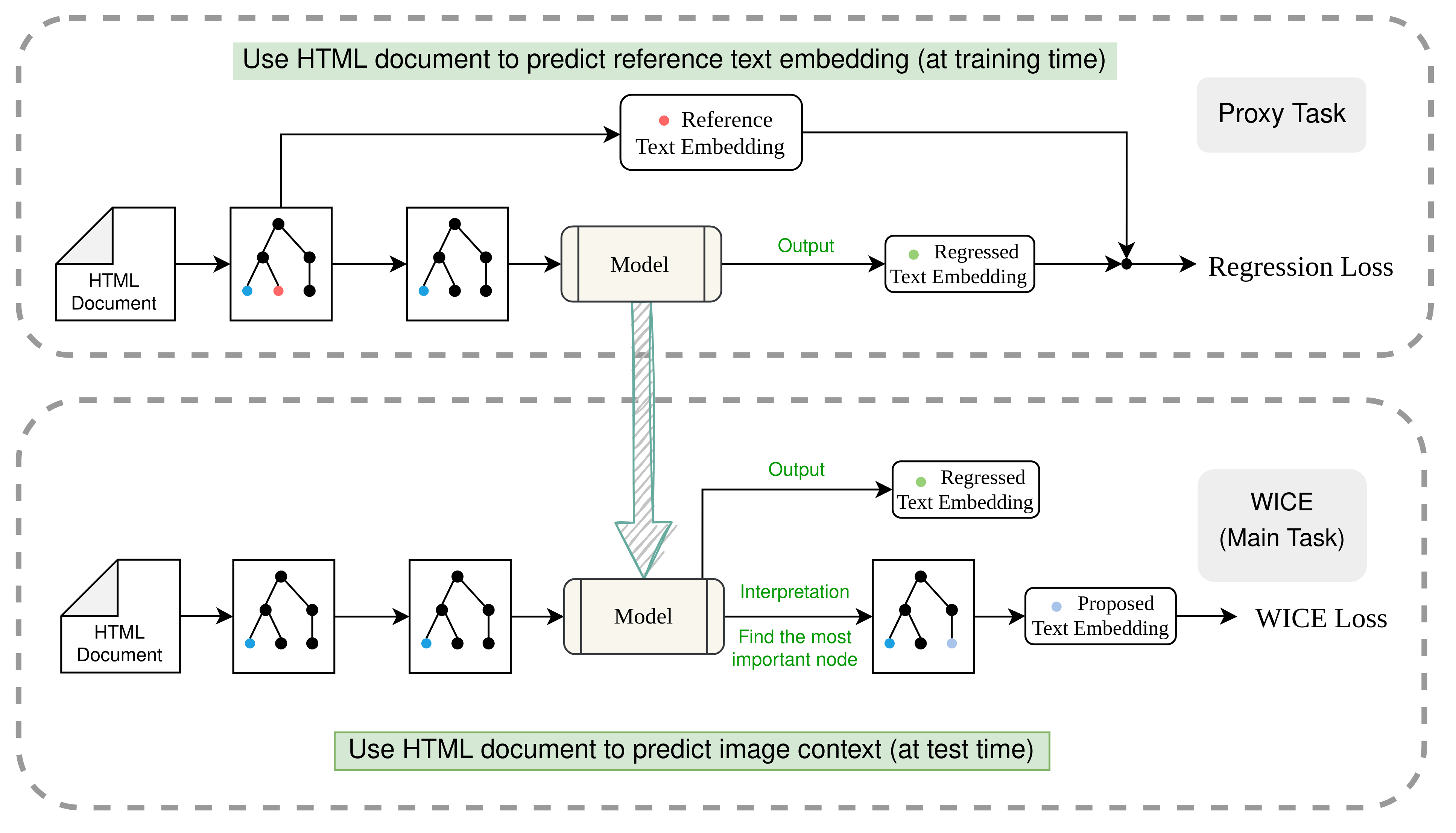}    
    \caption{Our pipeline. A graph neural network is first trained in the proxy task to predict the input HTML document's reference text (red dot), and then the importance score of the nodes (blue dots) in the prediction process is computed by interpreting the GCN model during the main task. The most predominant textual node (green dot) is then used as the context of the image. }

\label{fig:WICE_pipeline}
\end{figure}

\section{Method}

Our goal is to identify nodes in the DOM tree that may contain part of an image's context.
Since the DOM tree is a graph, we may use graph convolution networks (GCNs)~\cite{kipf2016semisupervised} to do so.
However, to bypass the lack of labeled datasets, we propose a {\em proxy task}, on unlabeled HTML documents to train our model. The results of this task may be interpreted to solve the WICE problem.
\cref{fig:WICE_pipeline} illustrates the sequence of steps in our method.

For each HTML document that contains an image, we use the longest text between the alternative text
\footnote{\mintinline{html}{"alt"} attribute of the \mintinline{html}{<img>} tag, which provides descriptive information of the image.}
, the caption
\footnote{text in \mintinline{html}{<figcaption>} that usually displays a short explanation besides the image.}
, and the image title
\footnote{\mintinline{html}{"title"} attribute of the \mintinline{html}{<img>} tag} as the {\em reference text}.
We assume here that the reference text always describes the image. This assumption may potentially cause bias.

While crawling news websites, we empirically found that approximately 50\% of them provide a caption or a reference text for their main illustrative images. This allows us to train on a relatively large corpus and also underlines the need for WICE in the wild: many websites do not provide clear textual contexts for their images.
\begin{figure}[t]
    \centering
    \includegraphics[width=0.9\textwidth]{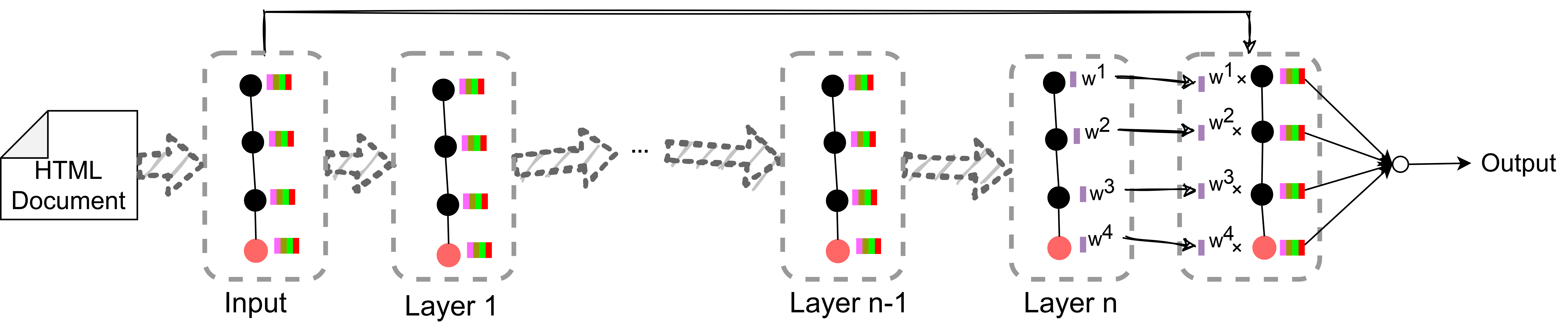}
    \caption{An illustration of the wGCN (weight-GCN) architecture. The nodes in the last layer have a feature dimension of 1. We interpret it as the importance score of each node. The output is the weighted average of the text embeddings of all nodes.}
    \label{fig:wGCN_illustration}
\end{figure}
Training models on a proxy task instead of directly predicting the context has many advantages. Firstly, we may rely on unlabeled datasets to train the model and solve the WICE problem, avoiding the need for annotations. It is easier and cheaper to crawl webpages with captions than manually annotate context sentences in thousands of articles. Moreover, at test time, our model will be used to infer the missing reference text: this means we can perform WICE even when the reference text is not present, which is where it is the most needed.

\subsection{Framework}

To extract the textual context of an image, we want to assign a weight $w_i$ to all text nodes $n_i$ from the DOM graph $\mathcal{G}$.
Since we cannot learn $w_i$ directly, we use a proxy task that consists of regressing a global embedding $\hat{\mathbf{z}}$ for the whole page.
We use as a supervised target the sentence embedding $\mathbf{z}^*$ of the reference text (alt-text or caption) extracted for the image.

For most of our models, we will assume that it is possible to reconstruct the reference text's embedding using a linear combination of the other text embeddings, \ie, we consider outputs in the form $\hat{\mathbf{z}} = \sum_i w_i f(\mathbf{z_i})$ where $\mathbf{z_i}$ is the embedding for the $i$-th node.
Since the regressed text embedding is obtained by averaging weighted node embeddings, we assume that the largest contributor, \ie the text node with maximum weight, is the most relevant for indexing.
The image context can thus be obtained by taking the text node $n_{\text{target}}$ that verifies $n_\text{target} = \argmax_i w_i$

In this work, the context is the most important text node, although this notion could be extended to all nodes over some threshold.

As a WICE metric for the extracted context, we compute the cosine similarity between the chosen text node's sentence embedding with the reference text's embedding, which is commonly used to measure document similarity in NLP.

\subsection{Baselines}

Inspired by the literature on WICE, we define various baselines for both the proxy and the WICE tasks.
We define the \texttt{distance} baseline, which weights a text node by the inverse of its distance $d_i$ to the image node in the graph:

\begin{align}
    \hat{\mathbf{z}} = \sum_i w_i \mathbf{z_i} = \sum_i \frac{1}{d_i} \cdot \mathbf{z_i}
\end{align}

%where $\mathbf{z_i}$ is the embedding for the $i$-th node, $d_i$ the distance in the graph between node $i$ and the image node, and $\hat{\mathbf{z}}$ the computed page embedding.

Implicitly, this baseline also defines an equivalent WICE baseline, which we call \texttt{text after image}, where the textual context of the image is the closest text node in the graph. This is known as "window-based context extraction" in the WICE literature \cite{SOUZA2004}.
As a simpler rule-based baseline, we also consider the \texttt{title} WICE heuristic that uses the \mintinline{html}{title} attribute of the webpage as context.

We define a \texttt{blind WICE} baseline as selecting the node with the most similar text (in the embedding space) across all text nodes, compared to the embedding regressed by the network ($w_i = 1$ if $i = \argmax_j (\cos(\hat{\mathbf{z}}, \mathbf{z_j}))$ and $0$ otherwise). Intuitively, it can be interpreted as ``looking for the missing caption''.
In this baseline, the WICE is ``blind'' since the reference text is completely unseen.
This defines a lower-bound for our method: as a sanity check, the most important node we find should at least describe the image better than the predicted embedding.

Finally, we define an \texttt{oracle} baseline that uses the actual reference text to find the most similar node. The resulting cosine similarity gives us an upper bound of what is achievable using our proxy model. It may be used as an indicator representing the model's performance potential: the smaller the gap between a model's results and the \texttt{oracle}, the better its performance.

\subsection{Models}

Two base models are studied to perform the text regression: Graph Convolution Network (GCN)~\cite{kipf2016semisupervised} and Graph Attention Network (GAT)~\cite{velikovi2017graph}.
GCNs are well-known for their promising performance on graph data.
However, it is difficult to explain the prediction because the fusion of graph structure and feature dimension achieved by GCN is an explicitly irreversible process \cite{xie2019interpreting}.
In comparison, GAT is a well-performing model, interpretable thanks to its attention mechanism which can be used as the weight of the text block.
Its multi-headed mechanism can also be utilized to stabilize the performance of the model.

We study two different approaches to help interpret the GCNs and produce the node weights vector $\mathbf{w}$.
First, we propose a GCN model that explicitly assigns weights to the nodes to facilitate the model's explanation, referred to as weight-GCN (wGCN).
A traditional GCN would map the entire graph $\mathcal{G}$ to the target embedding: $\mathbf{\hat{z}} = \Psi_\theta(\mathcal{G})$ where $\Psi$ is the GCN with parameters $\theta$.
However the information of which nodes contributed the most to the predictions is lost.
Instead, we make the GCN produce one weight per node. The regression result is then the weighted average embedding of all nodes in the graph.
\cref{fig:wGCN_illustration} illustrates the principle of the wGCN. Formally, let $\Psi_\theta$ be the GCN, $\mathbf{w} = \Psi_\theta(\mathcal{G})$ the output vector of node weights, and $\mathbf{z_i}$ be the text embedding of the node $i$ of graph $\mathcal{G}$. Then the regressed text embedding $\hat{\mathbf{z}}$ can be denoted as:

\begin{align}
 \hat{\mathbf{z}} = \sum_i \Psi_\theta(\mathcal{G})[i] \cdot \mathbf{z_i} = \sum_i w_i \mathbf{z_i}   
\end{align}

The wGCN is then trained using backpropagation and stochastic gradient descent: $\Psi_\theta^* = \argmin_\theta \mathcal{L}_\Psi(\theta)$. $\mathcal{L}$ is the proxy task loss function. We minimize the negative cosine similarity between $\hat{\mathbf{z}}$ and the reference text $\mathbf{z}^*$ \ie:
\begin{align}
\mathcal{L}_\Psi(\hat{\mathbf{z}}, \mathbf{z}^*) = 1 - \cos(\hat{\mathbf{z}}, \mathbf{z}^*) = 1 - \cos\left(\sum_i w_i\mathbf{z_i}, \mathbf{z}^*\right)
\end{align}

Our second approach uses the GAT attention scores as the weights $\mathbf{w}$.
The key difference between the GAT and wGCN is that wGCN learns the relationships between nodes and produces one weight per text node.
The regression embedding is then an average of the embeddings weighted by the wGCN scores.
In comparison, attention scores are only indirectly linked to the output embedding.

% We also considered using the Node Attribution Method (NAM) \cite{xie2019interpreting} to compute the weights $w_i$.
% However, since it is a gradient-based method, it would have required an additional backward pass which would have made it more expensive.%.% We did not use this approach in the end.
% to compute the importance of each node. It is a gradient-based method to calculate contribution of nodes in graph convolution networks.

We also used the DeeperGCN (DGCN) architecture \cite{li2019deepgcns,li2020deepergcn} to create deeper GCN models.
DGCN aims at solving common problems affecting deep GCNs, such as vanishing gradients, over-smoothing, and overfitting issues.
DGCN also uses recent deep learning tricks such as residual learning or dilated aggregation.

In our study, deeper neural networks make more sense because more neighbors can be explored.
Deeper networks lead to more nodes visited by the central node, and therefore more information collected.
This way, the image node receive information from all the article's nodes, even for pages with complicated DOM structures;
thus improving the representation capacities of the model.

\section{Experiments and results}

The dataset for our study was constructed using webpages from the Qwant News search index\footnote{\url{https://www.qwant.com/?t=news}}.
It consists of \num{242247} webpages from \num{1341} different websites, crawled mainly from French news websites.
Some Italian, German, and Spanish websites are also included.
Both international and regional websites from France are included in order to maximize diversity.

% % Option column wide, one below each other
% \begin{table}[ht]
% \caption{Optimal regression losses of each model}
% \label{tab:model_loss_comp}
% \input{tabularRegression.tex}
% % }
% \end{table}

% \begin{table}[ht]
% \caption{WICE cosine similarity loss.}
% \label{tab:WICE_loss}
% \input{tabularCosine.tex}
% \end{table}

We preprocess the HTML documents as follows: the content in \mintinline{html}{<main>}, \mintinline{html}{<body>} or \mintinline{html}{<article>} is first extracted from the webpage.
Some tags pertaining to layout, such as \mintinline{html}{<style>} or \mintinline{html}{<button>}, are then removed to clean up the DOM tree from unnecessary nodes.
Then, we extract the biggest image (in pixels) of a webpage with its reference text.
Webpages without such an image are removed from the dataset.
After preprocessing, the datasets contains \num{119550} webpages in \num{805} websites.
Texts are encoded using the multilingual sentence-BERT~\cite{reimers2019sentencebert,reimers2020making} that achieves state-of-the-art sentence embedding generation in several languages.
Node types are also considered useful and are one-hot encoded into 22 groups based on their HTML tag's semantics (lists, headers, paragraphs, etc.).
There are two ways to split the dataset: splitting per document regardless of the original website and split by website using all the pages of one website.
The second is more difficult because the data is not homogeneous: the test data may differ in structure and topic.
In the first setting, the ratio of webpages in the training set, validation set and test set are  5:2:3, \ie \num{59775} webpages in \num{723} websites for the training set, \num{23910} webpages in \num{622} websites for the validation set and \num{35865} webpages in \num{677} for the test set.
In the second setup, the proportions of websites in the training set, validation set and test set are also 5:2:3, \ie \num{53978} webpages from \num{402} websites in the training set, \num{28872} webpages from \num{162} websites in the validation set and \num{36700} webpages from \num{242} websites in the test set.
The optimal cosine similarity regression losses for each model of the two settings are shown in \cref{tab:model_loss_comp}.
As can be seen, the explicit weight-GCN model performs better on both settings and generalizes significantly better on unknown websites. We use only the wGCN architecture for the WICE task.

% Option column wide, one below each other
\begin{table}[t]
\caption{Proxy task regression performance for each model (average cosine similarity loss between the predicted embedding and the reference text embedding, lower is better).}
\label{tab:model_loss_comp}
\centering
\begin{tabularx}{0.9\textwidth}{ l YYYYYY}
 \toprule
			 & \multicolumn{3}{c}{\text{ split by webpages}} & \multicolumn{3}{c}{\text{ split by websites}} \\
                 \cmidrule(lr){2-4} \cmidrule(lr){5-7}
  Model &  ~train~ &  ~valid.~ & ~test~ & ~train~ & ~valid.~ & ~test~ \\
\midrule
\texttt{distance} & 0.587                                         & 0.589                       & 0.587                 & 0.618                  & 0.479                       & 0.609 \\
wGCN             &  \textbf{0.303}                                      &  \textbf{0.342}                    &  \textbf{0.339}              &  \textbf{0.357}               &  \textbf{0.355}                    &  \textbf{0.417} \\
GCN              & 0.365                                         & 0.407                       & 0.406                 & 0.426                  & 0.482                       & 0.569 \\
GAT              & 0.367                                         & 0.441                       & 0.439                 & 0.423                  & 0.563                       & 0.623 \\
DGCN             & 0.362                                         & 0.492                       & 0.493                 & 0.366                  & 0.618                       & 0.697 \\
\bottomrule
\end{tabularx}

% }
\end{table}

\begin{table}[t]
\caption{WICE performance (average cosine similarity loss between the context node embedding and the reference text embedding, lower is better).}
\label{tab:WICE_loss}
% \begin{tabular}{ @{} c *{6}{S} @{}}
% \toprule
%                  & \multicolumn{3}{c}{\text{ split by webpages}} & \multicolumn{3}{c}{\text{ split by websites}} \\
%                  \cmidrule(lr){2-4} \cmidrule(lr){5-7}
%   Model & \text{\quad train}                        & \text{\quad valid.} & \text{\quad~ test} & \text{\quad train} & \text{\quad valid.} & \text{\quad~ test} \\
% \midrule
% lower bound   & 0.293  & 0.297 & 0.293  & 0.334 & 0.264 & 0.259  \\
% \midrule
% wGCN   &  0.381 &  0.386  &  0.381 &  0.415 &  0.404  &  0.441  \\
% baseline & 0.654 & 0.658 & 0.653 & 0.715 & 0.508 & 0.680   \\
% random   & 0.780 & 0.779 & 0.779 & 0.792 & 0.736 & 0.800  \\
% \bottomrule
% \end{tabular}

\centering
\begin{tabularx}{0.9\textwidth}{ l YYYYYY}
\toprule
 & \multicolumn{3}{c}{\text{ split by webpages}} & \multicolumn{3}{c}{\text{ split by websites}} \\
 \cmidrule(lr){2-4} 
 \cmidrule(lr){5-7}
  Model & train &  {valid.} & {test} & train & valid. & test \\
\midrule
oracle   & 0.293  & 0.297 & 0.293  & 0.334 & 0.264 & 0.259  \\
\midrule
random   & 0.780 & 0.779 & 0.779 & 0.792 & 0.736 & 0.800  \\
title & 0.834 & 0.835  & 0.833  & 0.834 & 0.861 & 0.814   \\
text after image & 0.671 &  0.672 & 0.670 & 0.701 & 0.571 & 0.705 \\
blind WICE & 0.654 & 0.658 & 0.653 & 0.715 & 0.508 & 0.680   \\
wGCN   &  \textbf{0.381} &  \textbf{0.386}  &  \textbf{0.381} &  \textbf{0.415} &  \textbf{0.404}  &  \textbf{0.441}  \\
\bottomrule
\end{tabularx}
\end{table}

The average cosine similarity loss between the proposed text, which is the text with the highest score, and the image's reference text is shown in \cref{tab:WICE_loss}.
We see that naive WICE heuristics mostly fail on such a diverse dataset: \texttt{title} performs even worse than \texttt{random}, while \texttt{text after image} (which can be view as window-based WICE) rarely picks the best text node.
In theory, older works managed WICE using more complex heuristics.
However, defining a comprehensive ruleset does not scale up to more than \num{1000} websites and is unpractical in real applications.
We do not compare with visual-based WICE, either, because the rendering step using requires at least 1 second per webpage\footnote{With \eg, headless Chromium.}, \ie, more than three days for our whole dataset, not even including the segmentation algorithm.
\Chen{add the total execution time of wGCN?}
The preprocessing of our approach, \ie generating text embeddings and graphs, takes $\approx$ \num{0.429} second per webpage, totaling $\approx$ \num{21} hours.

For comparison, we add a random model that randomly choose a text node in the HTML page. Any model that has learnt anything should be better than this baseline.

The result shows that our model significantly outperforms WICE based on heuristics by at least \num{25}\%. The wGCN-extracted context is closer to the lower bound (\texttt{oracle}) than any other WICE approach we considered, thus validating our approach's relevance.
In practice, we found that a cosine similarity $\ge 0.6$ (\ie a cosine loss $\le 0.4$) is enough, considering that two sentences have the same topic. Our wGCN does not always reach this threshold in average but is significantly closer than other approaches that average around \num{0.3} similarity.

We found a strong correlation between regression losses and WICE losses during the exploration of the results (Pearson correlation coefficient: \num{0.96}), suggesting that better models for proxy tasks are better models for the main task.
We also found that texts with lower WICE losses are often semantically very close to the images and the reference texts or their topics.
This observation may be summarized as a correlation between lower regression loss and the relevance of the image for the extracted text (the original WICE problem's objective). This also shows that semantics may help in selecting nodes with similarly-named entities or dates, thus having a better chance to describe a given image.

We also observed that the model sometimes generalizes poorly to unknown websites, because of the heterogeneous webpages of each set.

However, this is less of a problem for closed set news crawler which parses a list of known websites regularly updated, with only the occasional introduction of a new domain.

\section{Conclusion}

In this work, we address the WICE problem by modeling and learning webpages using language models and GNNs, making large-scale automatic WICE easier, and not bounded by hard-coded rules.
We train a model on a large unlabeled news webpage corpus by learning to mimick the alt-text when it exists.
Our weight-GCN model assigns a weight to each text node and we then extract the most important node and define it as the image context.
This approach may blindly extract context sentences using semantic similarity between sentences and structural information learned from the DOM tree.
By working directly with the HTML, we avoid the need for rendering the webpage into an image and cut the preprocessing time by a factor 3, making large-scale WICE more tractable.

\bibliographystyle{splncs04}
\bibliography{sample-base}

\end{document}